\documentclass{article}

\usepackage[preprint]{neurips_2025}

\usepackage[utf8]{inputenc} % allow utf-8 input
\usepackage[T1]{fontenc}    % use 8-bit T1 fonts
\usepackage{graphicx}
\usepackage{hyperref}       % hyperlinks
\usepackage{url}            % simple URL typesetting
\usepackage{booktabs}       % professional-quality tables
\usepackage{amsfonts}       % blackboard math symbols
\usepackage{nicefrac}       % compact symbols for 1/2, etc.
\usepackage{microtype}      % microtypography
\usepackage{xcolor}         % colors
\usepackage{color,soul} % allows highlighting with the command "\hl{}"
\usepackage{amsmath}
\usepackage{xspace}

\usepackage{microtype}      % microtypography
\usepackage[most]{tcolorbox}
\usepackage{tikz}
\usepackage{amsmath}        % math symbols
\usepackage{multicol}       % multi-column layout inside boxes
\usepackage{array}
\usepackage{subcaption}
\usepackage{booktabs}
\usepackage{tabularx}
\usepackage{hyperref}
\DeclareMathOperator\arctanh{arctanh}

\newcommand{\hardmath}{\textbf{HARDMath2}\xspace}

\newcommand{\commentout}[1]{}
\newcommand{\thinline}{%
    \begin{tikzpicture}[baseline=-0.5ex]
        \draw[blue!25!black, line width=0.5pt, opacity=0.5] (0,0) -- (\linewidth,0);
    \end{tikzpicture}%
}

\title{HARDMath2: A Benchmark for Applied Mathematics Built by Students as Part of a Graduate Class}

\author{%
James V. Roggeveen\thanks{Equal contribution. Dataset available \href{https://huggingface.co/datasets/JVRoggeveen/HARDMath2}{here}.}, Erik Y. Wang\footnotemark[1], \\
Will Flintoft, Peter Donets, Lucy S. Nathwani, Nickholas Gutierrez, David Ettel, \\ 
Anton Marius Graf, Siddharth Dandavate,
Arjun Nageswaran, Raglan Ward,\\
Ava Williamson,  Anne Mykland, Kacper K. Migacz, Yijun Wang,  Egemen Bostan, \\
Duy Thuc Nguyen, Zhe He, Marc L. Descoteaux, Felix Yeung,
Shida Liu, \\
Jorge García Ponce, Luke Zhu, Yuyang Chen, Ekaterina S. Ivshina, 
Miguel Fernandez, \\
Minjae Kim, Kennan Gumbs, Matthew Scott Tan, Russell Yang,
Mai Hoang, \\
David Brown, Isabella A. Silveira, Lavon Sykes, Ahmed Roman, 
William Fredenberg, \\
Yiming Chen, Lucas Martin, Yixing Tang, Kelly Werker Smith, Hongyu Liao, \\
Logan G. Wilson, Alexander Dazhen Cai, Andrea Elizabeth Biju, Michael P. Brenner \\
\textbf{School of Engineering and Applied Sciences, Harvard University}
}

\begin{document}

\maketitle

\begin{abstract}
  Large language models (LLMs) have shown remarkable progress in mathematical problem-solving, but evaluation has largely focused on problems that have exact analytical solutions or involve formal proofs, often overlooking approximation-based problems ubiquitous in applied science and engineering. To fill this gap, we build on prior work and present \hardmath, a dataset of 211 original problems covering the core topics in an introductory graduate applied math class, including boundary-layer analysis, WKB methods, asymptotic solutions of nonlinear partial differential equations, and the asymptotics of oscillatory integrals. This dataset was designed and verified by the students and instructors of a core graduate applied mathematics course at Harvard. We build the dataset through a novel collaborative environment that challenges students to write and refine difficult problems consistent with the class syllabus, peer-validate solutions, test different models, and automatically check LLM-generated solutions against their own answers and  numerical ground truths. Evaluation results show that leading frontier models still struggle with many of the problems in the dataset, highlighting a gap in the mathematical reasoning skills of current LLMs. Importantly, students identified strategies to create increasingly difficult problems by interacting with the models and exploiting common failure modes. This back-and-forth with the models not only resulted in a richer and more challenging benchmark but also led to qualitative improvements in the students' understanding of the course material, which is increasingly important as we enter an age where state-of-the-art language models can solve many challenging problems across a wide domain of fields.
\end{abstract}

\section{Introduction} \label{sec:Introduction}

Recent advances in large language models (LLMs) have significantly expanded the frontier of automated mathematical reasoning. While early benchmarks largely focused on elementary arithmetic and symbolic algebra, newer datasets have begun to cover much more challenging material, ranging from Olympiad-style competition problems to graduate-level exams in theoretical mathematics. Indeed, recent benchmarks like \cite{glazer2024frontiermath} and \cite{phan2025humanity} include extremely challenging problems created by research mathematicians and  resist saturation even by today's most capable LLMs. However, a critical component of advanced mathematics crucial to applied science and engineering is severely  underrepresented. In many real-world contexts, equations that model physical systems, such as nonlinear partial differential equations (PDEs), oscillatory integrals, or multi-scale boundary-layer problems, do not admit exact solutions. Instead, analytical insights can be obtained from a sophisticated toolbox of asymptotic methods, perturbation expansions, and matched approximations. The ability to recognize and leverage these techniques is essential not only for human researchers but increasingly for AI systems intended to assist in scientific discovery. Existing benchmarks fail to capture this domain of reasoning in both scope and difficulty.

\hardmath was created to help address this gap,  consisting of 211 problems covering core topics from an introductory graduate course in applied mathematics. They consist of a mixture of original problems written by students enrolled in the course and questions adapted from standard textbooks \citep{bender2013advanced}. For the novel problems, students were asked to introduce complexities that would require careful reasoning and additional steps to solve. Each problem was solved by a student and peer-reviewed by other students to ensure the correctness of the solution, which served as the ground-truth against which the LLM-generated solutions were compared. Students revised their problems after seeing how the models responded, introducing additional facets that made the problem more difficult.

This collaborative approach led to a benchmark that is both diverse in content and challenging in form. It includes problems involving perturbation theory, nonlinear ordinary and partial differential equations (ODEs and PDEs), and challenging integrals. However, perhaps the most novel aspect of the dataset is how it was constructed. By interacting with LLMs during the problem-writing and problem-solving process, students simultaneously deepened their understanding of the material and were consequently able to write (and solve) problems that were more mathematically involved than the standard textbook problems posed on homework assignments. The interactive process was facilitated using a novel collaboration and evaluation environment in which students were able to contribute problems and improve on each other's work while being quickly graded based on evaluations from different LLMs. This dual perspective treated LLMs both as tools and as test subjects, pushing the students not only to understand the subtleties of solutions but also to craft and learn how to solve ever harder problems.  Our results show that even the most advanced models continue to struggle on many of our problems, as well as highlighting the educational value of building an LLM benchmark as part of a graduate class.

\section{Related work} \label{sec:related work}

Evaluations of mathematical reasoning have rapidly evolved alongside the models' capabilities. Early benchmarks played a crucial role in demonstrating the potential of LLMs in quantitative domains. Prominent examples include MATH \citep{hendrycks2021measuring}, comprising challenging high school competition-style problems, and GSM8K \citep{cobbe2021training}, which focused on multi-step arithmetic reasoning at the grade school level. However, many of these benchmarks are now saturated, giving rise to a new generation of benchmarks explicitly designed to test the limits of advanced mathematical reasoning. In many cases, these datasets have been curated by expert mathematicians and target difficulty levels comparable to graduate studies or mathematical research.

A notable example is \cite{glazer2024frontiermath}, which was developed through a collaboration involving over 60 mathematicians. The dataset contains original and unpublished problems spanning a wide range of modern pure mathematics, including number theory, algebraic geometry, category theory, and real analysis, and are touted to require hours or even days of effort from human experts. Another important effort is Humanity's Last Exam (HLE) \citep{phan2025humanity}, which aims to be a broad-coverage dataset at the frontiers of human knowledge. Unlike \cite{glazer2024frontiermath}, HLE was entirely crowdsourced from experts online, and consists of over 2,500 problems from a wide range of domains, with mathematics constituting just one topic. Other benchmarks have also targeted graduate-level material by sourcing problems directly from textbooks or qualifying examinations, such as GHOSTS \citep{frieder2024mathematical} (covering functional analysis, topology, and probability), ARB \citep{sawada2023arb} (consolidated from university mathematics qualifying exam problems), and s1-prob \citep{muennighoff2025s1simpletesttimescaling} (from the probability section of Stanford's PhD qualifying exam in statistics). While they contain challenging problems, these datasets are often limited in size and scalability, and tend to focus on abstract or formal mathematics. 

\subsection{The need for applied mathematics}

There is still a significant gap in coverage of advanced applied mathematics, where approximation methods allow insights to be gained from mathematical problems that might otherwise be intractable. Previous applied mathematics benchmarks, such as \cite{fan2025hardmath}, were limited by lack of diversity in problem types, phrasing templatization, and nonobjective method of evaluation. \hardmath addresses these limitations by introducing original problems that have unique forms using a student-driven method for data generation and an objective evaluation method. Finally, while some recent benchmarks \citep{feng2025physics, chung2025theoretical} target graduate-level physics problems, models will still struggle with such tasks if they cannot apply the prerequisite mathematical techniques covered in \hardmath. Moreover, if a model can solve advanced problems in physics but fails on the underlying math, it indicates that it is using faulty reasoning.

\subsection{Innovations in benchmark creation and pedagogical value}

Benchmarks are typically constructed using top-down, expert-driven design, as seen in FrontierMath and HLE \citep{glazer2024frontiermath, hendrycks2024humanity}, or via extraction from static sources like textbooks or exams \citep{frieder2024mathematical, sawada2023arb, petrov2025proof}. In contrast, our approach challenges students to design problems as they go through an applied mathematics course, focusing on problems tied to their current studies. We make use of an interactive environment to give students real-time feedback on the difficulty of their problems, which in turn allows them to iteratively increase the difficulty of their examples. Our methodology pushes students to interact with harder problems than typical for this course while fostering a course environment where an LLM becomes a tool for enriching education.

Finally, our evaluation framework allows objective assessment of the model's solution. Many mathematical benchmarks rely on LLM-based grading to evaluate the solution of a model. While this can capture the model's full solution---including its reasoning process---it introduces noise into the evaluation process, since the "grade" provided by an LLM depends highly upon the model and the prompt being used. Consequently, the LLM-as-a-judge approach to evaluation lacks reliability and objectivity. Other benchmarks use human experts to manually grade solutions according to a rubric. The drawback of this approach is the time required to assess the accuracy of a model and its labor-intensive nature. Our evaluations are conducted using an automatic parser that extract the final symbolic results of the model and ground-truth solutions, and compare them at a given point in the domain to determine whether they are numerically-close. While other benchmarks such as MathArena \citep{petrov2025proof} have also implemented automatic parsing of LaTeX solutions, we believe that our parser is so far the most sophisticated applied to mathematical benchmark grading.

% A notable advantage of this continuously expanding, student-driven dataset is its potential to enable longitudinal evaluation of frontier LLMs. By systematically tracking model performance across successive cohorts, this framework offers a unique opportunity to quantitatively assess the progression of LLM capabilities over time. Additionally, from an educational perspective, identifying the technical gaps in knowledge of frontier LLMs enables students to more deeply engage with the material.

\begin{table}
\centering
\small
\setlength{\tabcolsep}{4pt}
\caption{Comparison of \hardmath with selected advanced mathematical benchmarks. \hardmath distinctively targets graduate-level applied mathematics requiring approximations and features a student-driven, LLM-interactive problem creation process. Our evaluation method is also unique in that the final formula in the model's output is automatically compared against the ground-truth solution. }
\label{tab:hardmath2_comparison}
\resizebox{\textwidth}{!}{%
\begin{tabular}{@{}p{2.5cm}p{4.5cm}p{3.0cm}p{4.0cm}r@{}}
\toprule
Dataset & Math Focus & Problem Sourcing & Evaluation Method & Size \\
\midrule
FrontierMath  & Formal (Exact Solutions) & Expert Creation & Integer solution comparison & Hundreds \\
GHOSTS  & Formal (Proof-Based) & Manual Extraction & Manual human grading & 190 \\
ARB  & Formal (Proof-Based) & Manual Extraction & LLM-as-a-Judge & 34 \\
HLE  & Broad Coverage (Exact Solutions) & Crowd-Sourced & Multiple-choice comparison & ~1k \\
MathArena & Olympiad (Exact Solutions) & Expert Creation & Automated formula parsing & 96 \\
\midrule
\hardmath & Applied (Exact Solutions) & Student-Generated & Automated formula parsing & 211+ \\
\bottomrule
\end{tabular}

}
\end{table}

\section{Dataset and pedagogical framework for problem curation} \label{sec:dataset}

A distinctive aspect of this dataset is that it was created via the assignments of a university course. Students were tasked with designing at least one original problem each week that could not be solved correctly by Google's Gemini 2.0 Flash model, with an emphasis on creating problems in line with current topics being covered in the course. Students also had to provide a solution for their problems, which were required to be parsable by our custom evaluation framework and accompanied by a Colab notebook containing numerical verification of their approximate solution. Embedding the creation of the dataset into the core of the course turned homework assignments into a pedagogically rich and collaborative experience. LLMs became a tool to increase student engagement with challenging material, rather than a shortcut for bypassing it. The course culminated with an oral final exam for each student, where they were asked about a problem that they personally submitted to the benchmark. This ensured that the students understood the solutions to the problems they submitted, which were on average far more difficult than traditional homework problems for this course.

\subsection{Problem submission and verification pipeline} 

\begin{figure}
    \centering
    \includegraphics[width=0.95\linewidth]{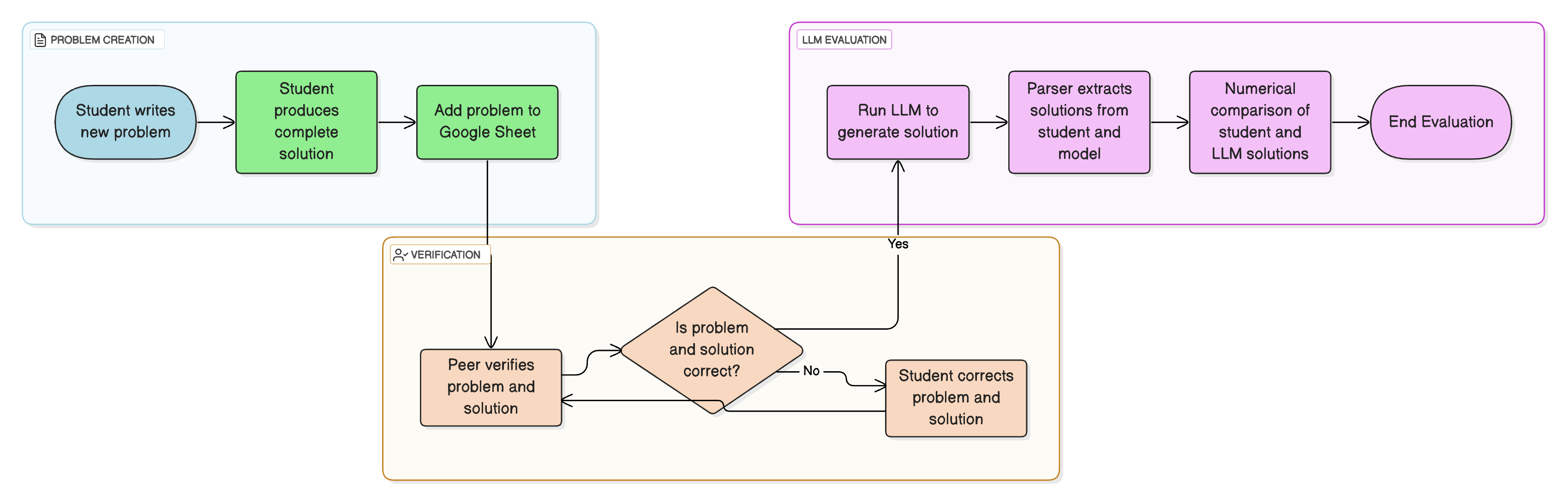}
    \caption{Flowchart of the problem-generation and validation process. Problem creation and validation happen on a collaborative Google Sheet, which includes custom functionality to send problems to a server for LLM querying and evaluation.}
    \label{fig:flowchart}
\end{figure}

Figure \ref{fig:flowchart} shows the pipeline for dataset creation. First, students write and solve new problems as part of their assigned coursework, and add their problems to a Google Sheet editable by all students in the course. To help facilitate this, the teaching staff gave standardized problem statements (discussed in Appendix \ref{A:Prompts}) that students could use to build their own questions. The students were then encouraged to edit these prompts as necessary to generate challenging questions. 

Student submissions were required to be parsable and gradable by our evaluation framework, which is discussed in detail in Section \ref{Section:infra}. This also meant that students had to carefully design the problem statement (which was given as the prompt to the LLM) to ensure that their problem clearly indicated a unique solution. One example of early trouble students had to rectify was that simply asking for an expansion without specifying the order leads to ambiguity in what an LLM might produce as a solution. Importantly, given that the student's solutions are the ground-truth for comparison against the LLM-generated solutions, they first went through collaborative verification. Students were required to submit with their problem a Google Colab notebook numerically verifying the accuracy of their approximate analytical solution. In addition to problem creation, students also had to verify each other's problems as part of their course assignment, using both the Colab notebook and whatever additional resources they wanted to use. These peer-reviewers were asked to correct any mistakes that were found either in the solutions or in the prompts to ensure the overall correctness of the dataset.  

Figure \ref{fig:sheet_validation} shows an example of the Google Sheet used for adding problems---which includes the original problem-designer, the verifier, the problem, solution, as well as checks on parsability and results from LLM evaluations---and an example of a validation plot generated by a student's Colab notebook. The final version of the problem and solution had to conform to a LaTeX format compatible with symbolic parsing, detailed in Appendix \ref{A:Evaluation}. However, students could request functionality be added to the parser when needed to increase the difficulty of a particular problem (one example was a student who requested support for incomplete Beta functions). 

In addition to our already-available dataset and evaluation code, we plan to publicly release the code to integrate LLM evaluation into Google Sheets, since we believe that this capability will be useful in the creation of future benchmarks. 

\subsubsection{Infrastructure for automating parsing and evaluation}\label{Section:infra}

\begin{figure}
    \centering
    \includegraphics[width=0.97\linewidth]{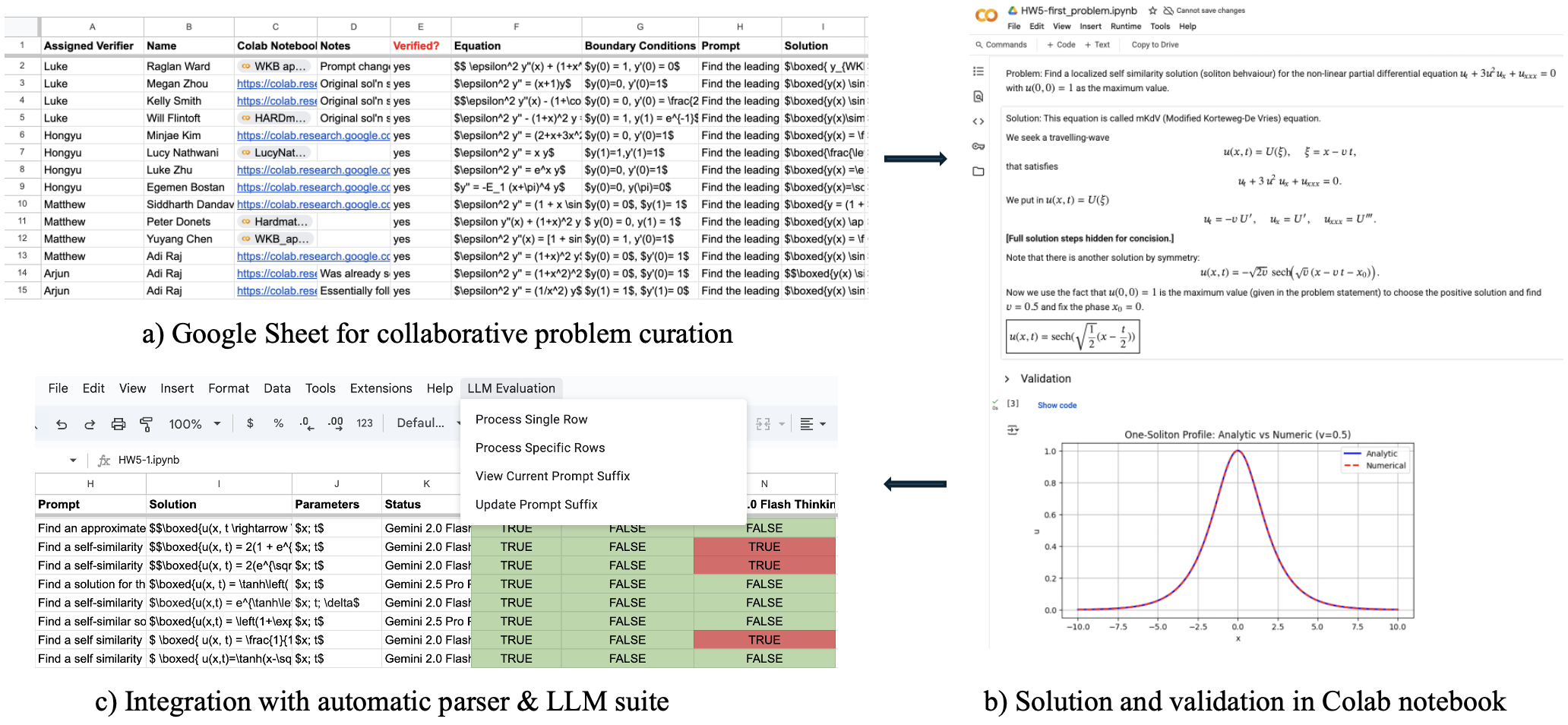}
    \caption{Problems are collected from students in a Google Sheet, which contains fields for all relevant aspects of the problem and solution, including the prompt passed to the LLM, the regime of interest, and additional parameters. Each student submitted a Colab notebook with their problem demonstrating a numerical comparison of their analytic solution to a full numerical solution, which could then be checked by student verifiers for accuracy. Then, students could instantly run an LLM on their problem (with standardized formatting and solution parsing automatically applied).}
    \label{fig:sheet_validation}
\end{figure}

We implemented a custom evaluator in Python that was hosted on an external server and interfaced with a custom Google Sheets plugin. This setup allowed students to write their problems, solutions, and any additional information (such as lists of the expected variables in the solution) directly in a spreadsheet. Students could then feed their problem directly from the spreadsheet to the evaluation server with the push of a button (Figure \ref{fig:sheet_validation}c). The parsing code first cleaned the raw LaTeX output by removing unnecessary formatting information before applying a series of regular expression replacement rules to transform LaTeX into a form that SymPy's \texttt{parse\_expr} function could interpret \citep{meurer2017sympy}. Some example transformations handled by our parser included rewriting expressions such as \texttt{\textbackslash sin\^{}2 } to \texttt{sin(x)\^{}2} and converting integrals like \texttt{\textbackslash int\_1\^{}2 e\^{}\{-\textbackslash frac\{3\}\{x\}\} dx} into \texttt{Integral(e**(-3/x), (x, 1, 2))}. We also included support for special functions that were commonly used in our dataset, such as the Gamma function and incomplete Beta functions. 

To compare two SymPy expressions, we numerically evaluated the expression by assigning a random value between one and two to every variable or parameter in the solution. This numerical verification step was necessary because the problems in our dataset may be solved using methods that produce the same formulae but could not be simply compared using SymPy's built-in equality checking. We required that all variables used in the solution be explicitly defined in the prompt and given to the evaluation code as a separate list, which forced problem-setters to write clear and unambiguous problems.

Coupling the parsing and evaluating code to a Google Sheet  allowed students to quickly receive feedback from the LLMs on their problems, letting them know if their problem was too easy or too ambiguous when the LLM's solution could not be correctly parsed. This collaborative format also enabled easy peer-verification and let us track how the difficulty of the benchmark evolved in real-time. The semi-automated solution verification and the standardization of prompt formats are discussed more thoroughly in Appendix~\ref{A:Evaluation}. The same parsing and evaluation code enabled on our Google Sheet was used to do the final evaluation of the models, as we made only a small subset of LLMs available to the students on the Google sheet. We found that both the problem authors and validators used the Google Sheet's access to the querying and evaluation framework to iteratively increase the difficulty of the problems and to provide a source of sanity-checking by comparing the LLMs reasoning to the student's solution.

\subsection{Problem types}

The problems in \hardmath cover many techniques from the applied mathematician's toolkit, such as the method of dominant balances, optimal truncation, boundary layer analysis, and asymptotic expansions. It goes significantly beyond \cite{fan2025hardmath}, which focused on more elementary topics.  A major distinction between \hardmath and other mathematical benchmarks is the combination of computational or numerical software, analytical techniques, and ``subjective" choices on the part of the problem-solver. For instance, to solve these problems, one must consider different regimes of solution space, the appropriate number of terms to include in approximate expressions, and which approximation method to use. These decisions are be made on a case-by-case basis but involve rigorous mathematical justification, and may be difficult tasks for existing LLMs.

\begin{figure}
    \centering
    \renewcommand{\arraystretch}{1.6}
    \footnotesize

    \begin{subfigure}[t]{0.48\textwidth}
        \centering
        \includegraphics[width=\linewidth]{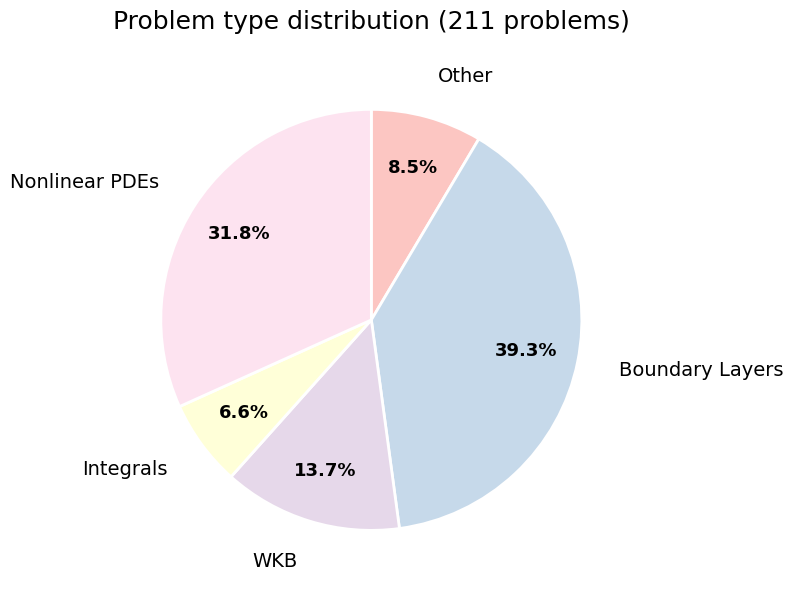}
        \caption{Distribution of problem types in \hardmath.}
        \label{fig:distribution}
    \end{subfigure}
    \hfill
    \begin{subfigure}[t]{0.48\textwidth}
        \centering
        \vspace*{-35mm} % Adjust this value to raise the table
        \resizebox{\linewidth}{!}{%
            \begin{tabular}{|>{\raggedright\arraybackslash}p{2.8cm}|
                            >{\raggedright\arraybackslash}p{3.0cm}|
                            >{\raggedright\arraybackslash}p{2.4cm}|}
            \hline
            \textbf{Problem type} & \textbf{Canonical form} & \textbf{Main tool} \\
            \hline
            Boundary layer &
            $\epsilon y'' - x y' + x^3 y = 0$, \newline
            $y(0) = A$, $y(1) = B$ &
            Matched \newline asymptotics \\
            \hline
            Traveling-wave PDE &
            $\partial_t u = D \partial_{xx} u + R(u)$ &
            Wave ansatz: \newline $u = f(x - vt)$ \\
            \hline
            WKB approximation &
            $\epsilon^2 y'' + Q(x) y = 0$ &
            WKB expansion: \newline $y \sim e^{S/\epsilon}$ \\
            \hline
            \end{tabular}
        }
        \caption{Canonical equation forms and their primary solution techniques.}
        \label{table:comparison}
    \end{subfigure}

    \caption{Problem type distribution and associated canonical solution forms in \hardmath.}
    \label{fig:dataset_combined}
\end{figure}

The dataset includes six distinct problem types that leverage these techniques: nonlinear PDEs, nonlinear ODEs, integrals, WKB approximations, boundary layers, and asymptotic series, with a distribution shown in Fig. \ref{fig:distribution}. As an example, boundary layer theory \citep{schlichting1961boundary} is an important applied mathematical tool that rectified apparent contradictions in the theory of aerodynamics. In the 1950s, the theory was further developed \citep{van1994nineteenth,lagerstrom2013matched} leading to both a widely-used toolkit for analyzing physical boundary value problems and a suite of canonically difficult problems that have challenged students for 75 years.
A design choice of \hardmath was to focus on problem types that can be made `harder' to solve (such as by introducing complicated forcing terms, as described in Section \ref{sec:design}), rather than simpler problem types that have limited complexity. Brief descriptions of the three largest problem classes are described below; the remaining problem types are discussed in detail in Appendix \ref{A:problems}.

% \begin{figure}
%     \centering
%     \includegraphics[width=0.5\linewidth]{Images/problem_breakdown2.png}
%     \caption{Distribution of problem types in the \hardmath dataset.}
%     \label{fig:dataset_distribution}
% \end{figure}

\subsubsection{Boundary layer problems}
\label{sec:blp}
Boundary‐layer problems arise in singularly perturbed differential equations where a small parameter $\epsilon$ multiplies the highest derivative. As $\epsilon \to 0$, solutions typically develop small regions with large gradients to satisfy the boundary conditions. An example of such a problem is given in Figure \ref{table:comparison}. In the limit of small $\epsilon$, the leading-order solution is determined by neglecting $\epsilon y''$, yielding $y_{\text{out}}$ but this solution fails to satisfy both boundary conditions. Therefore, one constructs \textit{inner solutions} by rescaling the independent variable near the boundaries to resolve the sharp gradients. The inner and outer solutions are then matched to form a uniformly valid solution.
% sounds good
% i wonder if we want to include a brief mathematical form for nonlinear pdes or boundary layer problems... primarily nonlinear pdes
% gotcha, maybe for nonlinear pdes though... i was going to d osomething like this:

\subsubsection{Nonlinear PDEs}
\label{sec:pdeprobs}
Several problems in the benchmark involve nonlinear PDEs, which feature terms where the solution or its derivatives appear nonlinearly; such problems in our benchmark can generally be written as
$$u_t + f(u, u_x, u_{xx}, \ldots) = 0,$$ where $f$ is a nonlinear function of the solution $u$ and its spatial derivatives. These equations can exhibit a range of behaviors depending on which terms in the PDE dominate. Diffusion-dominated solutions spread out over time, advection-dominated equations like Burgers’ equation generate shocks and discontinuities, dispersion-dominated systems like the Korteweg-de Vries equation produce solitons and wave trains, and solutions to Laplace's equation yield steady-state patterns or singularities. A given nonlinear PDE may exhibit different behaviors in different regions, depending on which terms dominate. Table~\ref{table:comparison} shows an example nonlinear PDE for which a traveling-wave ansatz simplifies the equation to an ODE.

% Several problems in the benchmark involve nonlinear PDEs that admit traveling-wave solutions. \hl{erik note: this example seems a bit too specific for this section} By applying the ansatz \( u(x,t) = f(x - vt) \), the PDE reduces to an ODE for the wave profile \(f(z)\). For equations of the form given in Table \ref{table:comparison}, this yields a second-order ODE that can often be integrated once to obtain a first-order equation in the phase plane. Boundary conditions at spatial infinity determine both the integration constant and the wave speed \(v\), typically through a heteroclinic connection.

\begin{comment}
    \hl{Siddharth Dandavate Portion} \hl{Al Liu; Streamlined and edited}
Assume \(u(x,t)=f(z)\) with \(z=x-vt\), so
\(\partial_tu=-v f'\), \(\partial_xu=f'\), \(\partial_x^2u=f''\), then the equation
\[
\partial_tu = D\,\partial_x^2u + R(u)
\]
which reduces to \(-\,v f' = D f'' + R(f).\) Integrating once gives the first‐integral \( D f' + v f + C = \int^{f}R(s)\,\mathrm{d}s,\) where the constant \(C\) is fixed by the limits \(f(-\infty)=u_-\) and \(f(+\infty)=u_+\).  These boundary conditions not only determine \(C\) but also select the wave speed \(v\) via a heteroclinic connection in the \((f,f')\) phase plane.  In the Fisher–KPP case \(R(f)=f(1-f)\), one finds explicit formulas for \(f(z)\) and the minimal speed \(v\); more generally, one applies phase‐plane, shooting, or perturbative methods to approximate the profile and velocity.

\hl{End Siddharth Dandavate Portion}
\end{comment}

\subsubsection{Wentzel–Kramers–Brillouin (WKB) approximations}
\label{sec:wkb}
We include linear ODEs that are solved with the WKB approximation. The WKB method approximates solutions to linear differential equations with a small parameter $\epsilon$ in the highest derivative term and is particularly effective in regimes where the solution varies rapidly. The solution is assumed to take the form $$y(x) \sim \exp\left( \frac{1}{\epsilon} S(x) \right),$$ where \(S(x)\) is expanded as an asymptotic series in $\epsilon$. Substituting into the differential equation and matching powers of $\epsilon$ yields a sequence of equations: the leading-order term \(S_0(x)\) satisfies a Hamilton–Jacobi-type equation, while higher-order terms correct the amplitude. The general solution is typically a linear combination of such exponential modes, matched to boundary conditions.

\section{Insights from students} \label{sec:design}
To create as challenging of a benchmark as possible, we asked students to design problems that frontier LLMs fail to solve. Using our interactive environment to test candidate problems and solutions, students identified consistent model weaknesses and created new problems that exploited these gaps. Three strategies used by students are as follows. 

\subsection{Structural obfuscation of canonical equations}

A common problem-solving technique used by both humans and LLMs involves matching equations to canonical forms. Consider the well-known Fisher–Kolmogorov–Petrovsky–Piskunov (Fisher–KPP) equation:
\begin{equation}
\frac{\partial u}{\partial t}
= D\,\frac{\partial^2 u}{\partial x^2}
\;+\;
r\,u\,(1 - u).
\label{eq:classical}
\end{equation}
By introducing an advection term and setting parameters, we can modify the PDE to
\begin{equation}
\frac{\partial u}{\partial t} - \frac{10}{\sqrt{30}} \frac{\partial u}{\partial x} = \frac{2}{5} \frac{\partial^2 u}{\partial x^2} + 2u(1 - u),
\label{eq:disguised}
\end{equation}
which is mathematically equivalent to the first form under a Galilean transformation. However, LLMs often fail to recognize the equivalence, as shown below.

\begin{tcolorbox}[breakable, colframe=blue!60!yellow, colback=blue!5!white, title={Error in Gemini 2.5 Pro Analysis}]
After transforming the PDE into the traveling wave ODE, the LLM states: "We eliminate the first derivative by choosing specific wave speed and therefore simplify the analysis, we set the coefficient of $ f' $ in equation (2) to zero. This eliminates the advection-like term in the ODE and reduces it to a conservative second-order form: $ c + \frac{10}{\sqrt{30}} = 0 \quad \Rightarrow \quad c = - \frac{10}{\sqrt{30}}. $"

{\color{red}Note: This arbitrary choice of $c$ is incorrect. For the specific ansatz $f(\xi) = \left( 1 + e^{a \xi} \right)^{-2}$ that the LLM subsequently (and correctly) proposed for the profile shape, the wave speed $c$ and the parameter $a$ are co-determined by the PDE's coefficients. This led to an incorrect reduced form and final solution.} 

% \\* \\*
% This incorrect assumption led the LLM to state: "Substituting into equation (2) gives:"$$f'' + 5 f(1 - f) = 0 $$
% The LLM then used the ansatz $f(\xi) = \left( 1 + e^{a \xi} \right)^{-2}$ and  substituted it into the simplified (but incorrectly derived for this ansatz) ODE.

% The LLM's incorrect final answer was $\boxed{u(x, t) = \left( 1 + \exp\left[\sqrt{\frac{5}{6}} \left(x + \frac{10}{\sqrt{30}} t\right) \right] \right)^{-2}}$.
\end{tcolorbox}

\commentout{
One can further obfuscate the PDE by embedding terms in abstract operators or using more complex functional forms. For example, the canonical KPP equation itself can be written as \(L[u] = 0\) where
\begin{equation}
L[u] = u_t - D\,u_{xx} - r\,u + r\,u^2.
\end{equation}

A more heavily disguised equation might involve replacing or augmenting standard terms with intricate expressions, such as an operator like
\begin{equation}
\mathcal{F}(u) = (1 + \alpha u + \beta u^2)\,\partial_x(\gamma\,u_x)
- \delta\,u_x\,u + \varepsilon\,u,
\end{equation}
which conceals simpler diffusion and reaction structures within composite algebraic and differential forms.}

\subsection{Introducing vanishing terms}

Differential equations can also be made more LLM-resistant simply by including complex nonlinear terms that evaluate to zero at the specific solution to the problem. For example, the function $
    u(x,t)=1+\tanh(e^{x-2t})$
satisfies the following PDE, which has a nonlinear term crafted to vanish at the solution:
$$\partial_{xx} u = \partial_x u + \frac{(\partial_t u)^2 (1 - u)}{2(2u - u^2)} + \underbrace{\frac{((\partial_{xx} u - \partial_x u)(2u - u^2) - \partial_x u \, \partial_t u (u - 1))}{((\partial_{tt} u)^2 + u^2)}}_{\text{additional term}}.
$$
Importantly, the PDE without the extra term at the end also has $u(x,t)$ as a solution and can often be solved by LLMs, but when confronted with the additional term, they incorrectly guess new functional forms that break their solution.

\subsection{Initial condition failures}
Even when LLMs identify correct general solutions, they often fail to apply initial conditions, which turn a general solution into a specific, physically meaningful one. For example, consider determining the leading-order behavior as \( x \to 0^+ \) of the third-order ODE:
\begin{equation}
x^4 \frac{d^3y}{dx^3} = y
\label{Eq:boundary_condition}
\end{equation}
with initial conditions $
y(1) = 1, y'(1) = 0, y''(1) = 0.$ A WKB expansion reveals exponential and oscillatory solutions. While models like Gemini 2.0 Flash Thinking and Gemini 2.5 Pro correctly identify this structure, they do not apply the initial conditions to combine the solutions. Even when prompted to do so, they return arbitrary values or keep the symbolic constants. The final answer from Gemini 2.5 Pro after being prompted to solve this equation is shown below:

\begin{tcolorbox}[breakable, colframe=blue!60!yellow, colback=blue!5!white, title={Gemini 2.5 Pro Output for Equation \ref{Eq:boundary_condition}}]
The leading order behavior of $y(x)$ as $x \to 0^+$ is given by
$y(x) \sim A x^{4/3} \exp(\frac{3}{2 x^{1/3}}) \cos(\frac{3\sqrt{3}}{2 x^{1/3}} - \phi)$
where $A$ is a positive constant, and $\phi$ is a phase constant. Since the constants $A$ and $\phi$ are fixed by the boundary conditions and are non-zero, this gives the leading behavior.

Final Answer: The final answer is $\boxed{y(x) \sim A x^{4/3} e^{3/(2x^{1/3})} \cos\left(\frac{3\sqrt{3}}{2x^{1/3}} - \phi\right)}$ for some constants $A>0$ and $\phi$.\newline

\color{red} Note: The LLM makes some attempt to narrow down what the constants are, but it does not solve for them completely. It mentions that the initial conditions fix the constants, but it does not list the equations it would use to solve for these constants, nor does it solve them.
\end{tcolorbox}

\section{Evaluation} \label{sec:evaluation}

To rigorously assess the abilities of current state-of-the-art language models, we conduct evaluations on a broad set of models. This includes closed-source models like OpenAI's GPT and o-series; Google's Gemini-series; and Anthropic's Sonnet 3.7, as well as open-source models such as Meta Llama 4 and DeepSeek V3. Closed-source models are accessed using their official APIs, while open-source models are tested via Together AI's API.

The results of our evaluation are presented in Table \ref{Table:eval_results}. We observe significant disparities across models and problem classes. Among the closed-source models, the Gemini 2.5 family---2.5 Flash Thinking and 2.5 Pro---exhibit the highest accuracies, achieving 60.1\% and 57.7\% respectively. These results show a marked improvement over prior Gemini models, which only attained 8.5\%. This suggests that the ``thinking process" built in to the newer Gemini models offer significantly improved mathematical reasoning capabilities. Similar improvements can be seen in OpenAI's o-series, which also demonstrate consistently strong performance. Notably, o3 achieves the highest overall score (52.5\%), followed closely by o4-mini (46.1\%) and o3-mini (46.0\%). In contrast, GPT-4.1 and GPT-4o show substantially worse performance. Even though GPT-4.1 was released relatively recently, its poorer performance may be attributed to its lack of a reasoning mode.

\begin{table}
\centering
\resizebox{\textwidth}{!}{
\begin{tabular}{|l||c|c|c|c|c|c|c|c|}
\hline
\textbf{Model} & \textbf{Overall} & \textbf{Asymptotic Series} & \textbf{Boundary Layers} & \textbf{Integrals} & \textbf{Nonlinear ODE} & \textbf{Nonlinear PDE} & \textbf{Other} & \textbf{WKB} \\
\hline
Claude 3.7 Sonnet & 11.8 & 33.3 & 4.2 & 61.5 & 28.6 & 1.7 & 16.7 & 24.0 \\
DeepSeek V3 & 18.9 & 0.0 & 22.5 & 46.2 & 42.9 & 6.7 & 0.0 & 18.2 \\
Llama 4 Maverick & 16.7 & 0.0 & 18.3 & 76.9 & 28.6 & 3.3 & 16.7 & 11.5 \\
GPT-4.1 & 7.4 & 0.0 & 1.4 & 38.5 & 16.7 & 5.1 & 0.0 & 14.3 \\
GPT-4o & 4.2 & 0.0 & 0.0 & 23.1 & 28.6 & 2.3 & 0.0 & 6.2 \\
o1 & 45.1 & 0.0 & 47.3 & 71.4 & 50.0 & 31.5 & 50.0 & 55.6 \\
o3 & 52.5 & 50.0 & 64.8 & 76.9 & 40.0 & 35.7 & 16.7 & 53.8 \\
o3-mini & 46.0 & 0.0 & 44.6 & 78.6 & 40.0 & 37.8 & 60.0 & 50.0 \\
o4-mini & 46.1 & 25.0 & 53.6 & 76.9 & 60.0 & 23.1 & 50.0 & 50.0 \\
Gemini 2.0 Flash & 8.5 & 0.0 & 1.3 & 33.3 & 33.3 & 9.3 & 0.0 & 13.6 \\
Gemini 2.5 Flash Thinking & 60.1 & 20.0 & 77.8 & 71.4 & 40.0 & 45.0 & 16.7 & 61.5 \\
Gemini 2.5 Pro Preview & 57.7 & 25.0 & 72.6 & 78.6 & 14.3 & 44.3 & 20.0 & 60.0 \\
\hline
\end{tabular}
}
\caption{Pass@1 Rates by Model and Question Type}
\label{Table:eval_results}
\end{table}

Within the dataset, integrals appear to be the most tractable problem type, with models like Llama 4 Maverick, o1, o3, o3-mini, and Gemini 2.5 Pro Preview all scoring above 70\%. This suggests that problems that must conform to a relatively rigid structure may be more easily solved by state-of-the-art models. Nonlinear PDEs, by contrast, are considerably more challenging; only the top-performing models exceed 35\% in this category, with o3-mini and Gemini 2.5 variants reaching 37.8\% and 44.3–45.0\%, respectively. This provides further evidence that problem types which can be more substantially customized---especially via the techniques discussed in Section \ref{sec:design}---pose a greater challenge to models. Finally, some models struggled with instruction-following, such as failing to box their final solutions or using improper formatting. This is discussed in Appendix \ref{Sec:model_analysis}.

\begin{figure}
    \centering
    \begin{subfigure}[t]{0.41\textwidth}
        \centering
        \includegraphics[width=\textwidth]{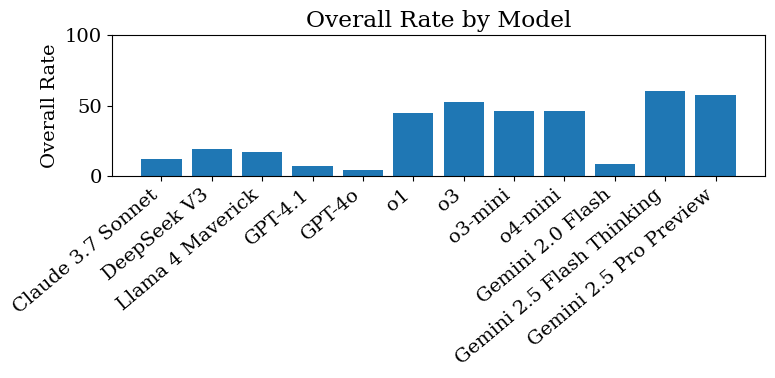}
        \caption{Overall performance of models across all problem types in \hardmath.}
        \label{fig:overall_performance}
    \end{subfigure}
    \begin{subfigure}[t]{0.56\textwidth}
        \centering
        \raisebox{1.5pt}{
        \includegraphics[width=\textwidth]{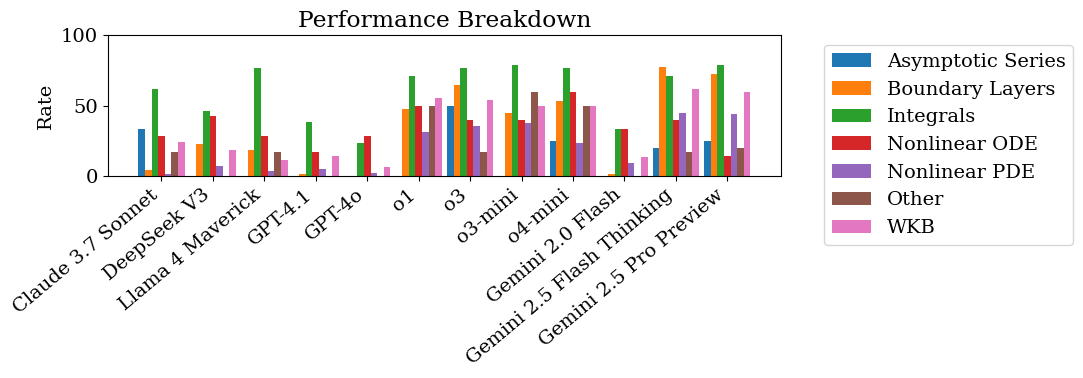}}
        \caption{Breakdown of model performance across problem types.}
        \label{fig:performance_breakdown}
    \end{subfigure}
    \caption{Model performance on \hardmath. While (a) shows overall success rates by model, (b) shows significant differences in performance across problem types.}
    \label{fig:combined}
\end{figure}

Overall, the results in Figure \ref{fig:overall_performance} show that the benchmark remains difficult for even the most capable LLMs currently available. While top-performing models like Gemini 2.5 Pro and o3 demonstrate competence in some categories, no model approaches mastery across all problem types. In categories such as boundary layer theory, nonlinear PDEs, and WKB approximation, the majority of models fail to achieve even moderate accuracy. The consistently low scores outside integrals and basic ODEs highlight persistent gaps in multi-step mathematical reasoning and problem-solving techniques that are fundamental to advanced applied mathematics. These results therefore suggest that while recent advancements in LLMs have advanced the frontier in terms of their mathematical abilities, the specialized mathematical knowledge required by this benchmark still remains largely out of reach.

% \begin{figure}
%     \centering
%     \includegraphics[width=0.85\linewidth]{Images/performance_breakdown2.png}
%     \caption{Accuracy of language models across problem categories. Bars show category-wise performance, while the black dots show overall accuracy across all problem types per model.}
%     \label{fig:enter-label}
% \end{figure}

\section{Conclusion} 
We introduce \hardmath, a benchmark that is inspired by previous work \citep{fan2025hardmath} but contains a variety of harder problem types from a graduate applied mathematics course. The benchmark is unique in both scope and design, comprising 211 original problems across several categories: nonlinear PDEs, integrals, WKB approximations, boundary layer problems, and asymptotic expansions. It expands the current landscape of mathematical benchmarks by introducing more challenging and underrepresented applied mathematics problems involving techniques crucial to real-world scientific and engineering applications. Notably, it also features a novel approach to problem curation that leverages modern LLMs to enhance student learning.

Going forward, we plan to broaden the range of problem types in the benchmark, focusing on the problem types that can be made more difficult. Examples of such problems include asking for more terms in an asymptotic expansion or solving other types of PDEs, like Green’s functions. Students similarly found that LLMs themselves could be used to arbitrarily make problems more difficult via prompting, although this would also require more time and effort from the students due to the generation-verification gap \citep{songmind}. Finally, our hybrid student and LLM-driven framework for designing original problems that are more challenging than those found in textbooks or assigned on homework assignments can be applied to any quantitative class involving advanced mathematical reasoning, such as courses in physics, statistics, and engineering. 

\label{sec:conclusion}

\bibliographystyle{plainnat}
\bibliography{bib}

%%%%%%%%%%%%%%%%%%%%%%%%%%%%%%%%%%%%%%%%%%%%%%%%%%%%%%%%%%%%

% \appendix

\newpage

\appendix
\section{Appendix} \label{sec:appendix}

\subsection{Detailed descriptions of problem types} \label{A:problems}

\subsubsection{Boundary layer problems} \label{A:Boundary Layer}
Many problems in \hardmath are ODEs with boundary layers, where different solutions dominate near a boundary than in the rest of the region (see Section~\ref{sec:blp}). We provide an example below.
\begin{tcolorbox}[breakable, colframe=blue!60!yellow, colback=blue!5!white, title=Sample Boundary Layer Problem and Full Solution]
\textbf{Problem:} 
Consider the differential equation:
\[
\epsilon y''-xy+x^3y=0
\]
Find a uniformly valid approximation to the solution with boundary conditions $y(0)=A$ and $y(1)=B$ in the limit $\epsilon\ll 1$. 

\thinline

\textbf{Solution:}

Step 1: Find the outer solution
Use dominant balance since $\epsilon$ is small:
\[
-x y_{out}' + x^3 y_{out} = 0
\]
If $x \neq 0$, this simplifies to $y_{out}' = x^2 y_{out}$.This is a separable first-order differential equation:
\[\frac{dy_{out}}{y_{out}} = x^2 dx\]
Integrating both sides:
\[
\ln|y_{out}| = \frac{x^3}{3} + C_0
\]
\[
y_{out}(x) = C e^{x^3/3}
\]
where $C$ is an arbitrary constant.

Step 2: Analyze boundary conditions and potential boundary layers.

The outer solution must satisfy one of the boundary conditions. Check $y(0)=A$.
$y_{out}(0) = C e^0 = C$. So, $C=A$.Thus, the outer solution satisfying $y(0)=A$ is:
\[y_{out}(x) = A e^{x^3/3}\]

Check the behavior of this outer solution at $x=0$ with respect to the original equation.
\[y_{out}'(x) = A x^2 e^{x^3/3}\]
\[y_{out}''(x) = A (2x e^{x^3/3} + x^2 \cdot x^2 e^{x^3/3}) = A (2x + x^4) e^{x^3/3}\]

Substituting these into the full differential equation at $x=0$:
\[\epsilon y_{out}''(0) - 0 \cdot y_{out}'(0) + 0^3 \cdot y_{out}(0) = \epsilon \cdot 0 - 0 + 0 = 0\]
The outer solution $y_{out}(x) = A e^{x^3/3}$ satisfies the boundary condition at $x=0$ and is consistent with the full differential equation at $x=0$. This suggests there is no boundary layer at $x=0$.

Consider the boundary condition at $x=1$, $y(1)=B$.
The outer solution gives $y_{out}(1) = A e^{1/3}$.
If $B \neq A e^{1/3}$ (which is generally the case), the outer solution alone cannot satisfy the boundary condition at $x=1$. This indicates a boundary layer near $x=1$.

Step 3: Analyze the boundary layer at $x=1$.

Let the inner variable be $X = \frac{1-x}{\epsilon^\alpha}$ for some $\alpha > 0$. So $x = 1 - \epsilon^\alpha X$. Thus, the derivatives become:
\[
y' = -\frac{1}{\epsilon^\alpha} \frac{dy}{dX}
\]
\[
y'' = \frac{1}{\epsilon^{2\alpha}} \frac{d^2y}{dX^2}
\]
Substituting these into the differential equation:
\[
\epsilon^{1-2\alpha} \frac{d^2y}{dX^2} + \epsilon^{-\alpha}(1-\epsilon^\alpha X) \frac{dy}{dX} + (1-3\epsilon^\alpha X + O(\epsilon^{2\alpha})) y = 0
\]
We need to balance the highest derivative term with another dominant term in the boundary layer. The term $\epsilon^{-\alpha} \frac{dy}{dX}$ is likely to be dominant. So, $1-2\alpha = -\alpha \implies \alpha=1$

The stretched variable is $X = \frac{1-x}{\epsilon}$.
The equation in terms of $X$ for the inner solution $y_{in,1}(X)$ becomes (substituting $\alpha=1$):
\[
\frac{d^2y_{in,1}}{dX^2} + (1-\epsilon X) \frac{dy_{in,1}}{dX} + \epsilon (1-\epsilon X)^3 y_{in,1} = 0
\]
To leading order in $\epsilon$, as $\epsilon \to 0$:
\[
\frac{d^2y_{in,1}}{dX^2} + \frac{dy_{in,1}}{dX} = 0
\]
The general solution is $y_{in,1}(X) = K_1 e^{-X} + K_2$

Step 4: Apply boundary condition and matching for the inner solution at $x=1$.

The boundary condition at $x=1$ is $y(1)=B$. In terms of $X$, $x=1$ corresponds to $X=0$. So:
\[
y_{in,1}(0) = B \implies K_1 e^0 + K_2 = B \implies K_1 + K_2 = B
\]
Matching condition: The limit of the inner solution as $X \to \infty$ must match the limit of the outer solution as $x \to 1^-$.
\[
\lim_{X \to \infty} y_{in,1}(X) = \lim_{X \to \infty} (K_1 e^{-X} + K_2) = K_2
\]
\[
\lim_{x \to 1^-} y_{out}(x) = \lim_{x \to 1^-} A e^{x^3/3} = A e^{1/3}
\]
\[
K_2 = A e^{1/3}
\]
Substituting $K_2$ back into $K_1+K_2=B$:
\[
K_1 + A e^{1/3} = B \implies K_1 = B - A e^{1/3}
\]
Thus, the inner solution for the boundary layer at $x=1$ is:
\[
y_{in,1}\left(\frac{1-x}{\epsilon}\right) = (B - A e^{1/3})e^{-(1-x)/\epsilon} + A e^{1/3}
\]
Step 5: Construct the uniformly valid approximation.
A common method for constructing a uniformly valid approximation is $y_{unif}(x) = y_{out}(x) + y_{in}(x) - \text{common part}$. So we have:
\[
y_{unif}(x) = A e^{x^3/3} + (B - A e^{1/3})e^{-(1-x)/\epsilon}
\]

Step 6: Verify boundary conditions for the uniform approximation.

At $x=0$:
\[
y_{unif}(0) = A e^0 + (B - A e^{1/3})e^{-1/\epsilon} = A + (B - A e^{1/3})e^{-1/\epsilon}
\]
Since $\epsilon \ll 1$, $e^{-1/\epsilon}$ is exponentially small. So $y_{unif}(0) \approx A$.

At $x=1$:
\[
y_{unif}(1) = A e^{1/3} + (B - A e^{1/3})e^0 = A e^{1/3} + B - A e^{1/3} = B
\]

The boundary conditions are satisfied to leading order.
The final solution is:
\[\boxed{y(x) \approx A e^{x^3/3} + (B - A e^{1/3})e^{-(1-x)/\epsilon}}\]
\end{tcolorbox}

\subsubsection{Nonlinear PDE problems} \label{A:Boundary Layer}
We solve many types of nonlinear PDEs in the benchmark (Section~\ref{sec:pdeprobs}). Here, we provide a solution to a PDE dominated by dispersion with a traveling wave solution. 
\begin{tcolorbox}[breakable, colframe=blue!60!yellow, colback=blue!5!white, title=Sample Nonlinear PDE Problem and Full Solution]
\textbf{Problem:} 
Consider the Korteweg-de Vries (KdV) equation:
\[
\frac{\partial u}{\partial t} + 6 u \frac{\partial u}{\partial x} + \frac{\partial^3 u}{\partial x^3} = 0.
\]
Find a soliton solution in the limit $t\Rightarrow\infty$.

\thinline

\textbf{Solution:} 
We seek a traveling wave solution of the form $u(x, t) = f(\xi)$, where $\xi = x - c t$ and $c$ is the constant wave speed. The domain is approximated as $x \in (-\infty, \infty)$ for a localized soliton solution.
Substituting the traveling wave ansatz into the KdV equation yields:
\[
-c f'(\xi) + 6 f(\xi) f'(\xi) + f'''(\xi) = 0
\]
Integrating once with respect to $\xi$:
\[
f'' + 3 f^2 - c f + A = 0
\]
For a localized solution, we require $f, f', f'' \to 0$ as $|\xi| \to \infty$. This boundary condition implies the integration constant $A=0$.
\[
f'' + 3 f^2 - c f = 0
\]
Multiplying by $f'$ to facilitate integration (energy method):
\[
f' f'' + 3 f^2 f' - c f f' = 0
\]
This can be written as the derivative of a conserved quantity:
\[
\frac{d}{d \xi} \left( \frac{1}{2} (f')^2 + f^3 - \frac{c}{2} f^2 \right) = 0
\]
Integrating again with respect to $\xi$:
\[
\frac{1}{2} (f')^2 + f^3 - \frac{c}{2} f^2 + B = 0
\]
Applying the boundary conditions $f, f' \to 0$ as $|\xi| \to \infty$ requires the second integration constant $B=0$.
\[
\frac{1}{2} (f')^2 = \frac{c}{2} f^2 - f^3
\]
Rearranging gives:
\[
(f')^2 = c f^2 - 2 f^3 = f^2 (c - 2f)
\]
Assuming $f>0$ within the soliton and taking the square root ($f' = \frac{df}{d\xi}$):
\[
\frac{df}{d\xi} = \pm f \sqrt{c - 2f}
\]
Separating variables:
\[
\frac{df}{f \sqrt{c - 2f}} = \pm d\xi
\]
Integrating both sides:
\[
\int \frac{df}{f \sqrt{c - 2f}} = \pm \int d\xi = \pm (\xi - \xi_0)
\]
where $\xi_0$ is an integration constant representing the initial position.

To evaluate the integral on the left, we use the substitution $f = \frac{c}{2} \text{sech}^2(\theta)$. 
\[
df = -c \, \text{sech}^2(\theta) \tanh(\theta) \, d\theta
\]
The term under the square root becomes:
\[
\sqrt{c - 2f} = \sqrt{c - c \, \text{sech}^2(\theta)} = \sqrt{c (1-\text{sech}^2(\theta))} = \sqrt{c \tanh^2(\theta)} = \sqrt{c} |\tanh(\theta)|
\]
Choose the branch where $\tanh(\theta) > 0$:
\[
\int \frac{-c \, \text{sech}^2(\theta) \tanh(\theta) \, d\theta}{\left(\frac{c}{2} \text{sech}^2(\theta)\right) (\sqrt{c} \tanh(\theta))} = \int \frac{-2}{\sqrt{c}} d\theta = -\frac{2}{\sqrt{c}} \theta
\]
Equating this to the right side:
\[
-\frac{2}{\sqrt{c}} \theta = \pm (\xi - \xi_0)
\]
\[
\theta = \mp \frac{\sqrt{c}}{2} (\xi - \xi_0)
\]
Substituting back into $f = \frac{c}{2} \text{sech}^2(\theta)$:
\[
f(\xi) = \frac{c}{2} \text{sech}^2\left( \mp \frac{\sqrt{c}}{2} (\xi - \xi_0) \right)
\]
Finally, substituting $\xi = x - ct$, and set $c = 4$ $x_0=0$:
\[
\boxed{u(x,t)=2 \text{sech}^2(x - 4 t) }
\]

\end{tcolorbox}

\subsubsection{WKB approximation problems} \label{A:wkb}
We include many ODEs that can be modeled using the WKB approximation; see Section \ref{sec:wkb} for an explanation of the technique. We provide a simple example problem below.

\begin{tcolorbox}[breakable, colframe=blue!60!yellow, colback=blue!5!white, title=Sample WKB Problem and Full Solution with Initial Conditions]
\textbf{Problem:}
Consider the differential equation:
\[
y''(x) = \frac{x}{\epsilon^2} y(x),
\]
for small positive $\epsilon$ in the limit $\epsilon\ll 1$, subject to the initial conditions at $x=1$:
\[
y(1) = e^{2/(3\epsilon)}, \quad y'(1) = \frac{1}{\epsilon} e^{2/(3\epsilon)}.
\]

\thinline

\textbf{Solution:} This equation fits the general WKB form:
\[
y'' = R(x)y \quad \text{with} \quad R(x) = \frac{x}{\epsilon^2}.
\]
We assume a solution of the form:
\[
y(x) \sim \exp\left(\frac{1}{\delta} \sum_{n=0}^\infty \delta^n S_n(x)\right).
\]
To leading order, we approximate this by truncating after the first two terms:
\[
y(x) \sim \exp\left(\frac{1}{\delta}(S_0(x) + \delta S_1(x))\right).
\]
We now differentiate using the product rule:
\[
y' = \left(\frac{1}{\delta} S'(x)\right) \exp\left(\frac{1}{\delta} S(x)\right), \quad
y'' = \left[\left(\frac{1}{\delta} S'(x)\right)^2 + \left(\frac{1}{\delta} S''(x)\right)\right] \exp\left(\frac{1}{\delta} S(x)\right).
\]
Substitute into the original differential equation:
\[
\left(\frac{1}{\delta} S'(x)\right)^2 + \left(\frac{1}{\delta} S''(x)\right) = \frac{x}{\epsilon^2}.
\]
Expanding and collecting powers of $\delta$ gives:
\[
\delta^{-2} S_0'^2 + 2\delta^{-1} S_0' S_1' + \delta^{-1} S_0'' + \cdots = \frac{x}{\epsilon^2}.
\]
The leading-order balance suggests that $S_0'^2 \sim x$, and we expect $S_0'$ to be large while $S_0''$ remains small. Thus, to leading order, we take:
\[
\delta^{-2} S_0'^2 = \frac{x \delta^2}{\epsilon^2}.
\]
To match both sides, we must take $\delta = \epsilon$, the small parameter. Substituting back in:
\[
S_0'(x)^2 = x \quad \Rightarrow \quad S_0(x) = \pm \int_0^x \sqrt{t} \, dt = \pm \frac{2}{3} x^{3/2}.
\]

Now solve for the first-order correction $S_1(x)$. From the remaining terms:
\[
2 S_0' S_1' + S_0'' = 0.
\]
Using $S_0' = \sqrt{x}$ and $S_0'' = \frac{1}{2\sqrt{x}}$, we get:
\[
2 \sqrt{x} S_1' + \frac{1}{2\sqrt{x}} = 0 \quad \Rightarrow \quad S_1'(x) = -\frac{1}{4x}, \quad S_1(x) = -\frac{1}{4} \ln x.
\]

Combining these, we find the two independent asymptotic solutions:
\[
y_1(x) \sim x^{-1/4} \exp\left(\frac{2 x^{3/2}}{3\epsilon}\right), \quad
y_2(x) \sim x^{-1/4} \exp\left(-\frac{2 x^{3/2}}{3\epsilon}\right)
\]

These represent the two dominant behaviors of the solution in the limit $\epsilon \to 0$. The exponential terms capture rapid growth or decay, while the $x^{-1/4}$ prefactor corrects the amplitude to leading order. The general solution is a linear combination of these modes that satisfies the boundary conditions.

The general solution is $y(x) \approx c_1 y_1(x) + c_2 y_2(x)$. We apply the initial conditions at $x=1$.
Using the WKB solutions at $x=1$:

$y_1(1) = 1^{-1/4} \exp\left(\frac{2 (1)^{3/2}}{3\epsilon}\right) = e^{2/(3\epsilon)}$, and $y_2(1) = 1^{-1/4} \exp\left(-\frac{2 (1)^{3/2}}{3\epsilon}\right) = e^{-2/(3\epsilon)}$

Using the leading-order WKB derivative approximation $y'(x) \approx \frac{S_0'(x)}{\epsilon} y(x) = \frac{\sqrt{x}}{\epsilon} y(x)$:

$$y_1'(1) \approx \frac{\sqrt{1}}{\epsilon} y_1(1) = \frac{1}{\epsilon} e^{2/(3\epsilon)}$$

$$y_2'(1) \approx -\frac{\sqrt{1}}{\epsilon} y_2(1) = -\frac{1}{\epsilon} e^{-2/(3\epsilon)}$$

Applying the initial conditions $y(1) = e^{2/(3\epsilon)}$ and $y'(1) = \frac{1}{\epsilon} e^{2/(3\epsilon)}$:
\begin{align*} y(1) &= c_1 y_1(1) + c_2 y_2(1) \\ e^{2/(3\epsilon)} &= c_1 e^{2/(3\epsilon)} + c_2 e^{-2/(3\epsilon)} \end{align*}
Divide the first equation by $e^{2/(3\epsilon)}$:
\[ 1 = c_1 + c_2 e^{-4/(3\epsilon)} \]
Applying the second initial condition:
\begin{align*} y'(1) &= c_1 y_1'(1) + c_2 y_2'(1) \\ \frac{1}{\epsilon} e^{2/(3\epsilon)} &\approx c_1 \left(\frac{1}{\epsilon} e^{2/(3\epsilon)}\right) + c_2 \left(-\frac{1}{\epsilon} e^{-2/(3\epsilon)}\right) \end{align*}
Divide the second equation by $\frac{1}{\epsilon} e^{2/(3\epsilon)}$:
\[ 1 \approx c_1 - c_2 e^{-4/(3\epsilon)} \]
For small $\epsilon$, $e^{-4/(3\epsilon)}$ is extremely small. To leading order in $\epsilon$:
$1 = c_1 + c_2 \cdot (\text{very small number}) \implies c_1 \approx 1$
$1 = c_1 - c_2 \cdot (\text{very small number}) \implies c_1 \approx 1$
Substituting $c_1 \approx 1$ into the first equation gives $1 = 1 + c_2 e^{-4/(3\epsilon)}$, which implies $c_2 e^{-4/(3\epsilon)} = 0$. Since $e^{-4/(3\epsilon)} \neq 0$, we must have $c_2 = 0$.

Thus, these specific initial conditions select $c_1 \approx 1$ and $c_2 = 0$. The resulting solution is approximately $y(x) \sim y_1(x)$.

The specific solution satisfying these initial conditions is therefore the positive exponential branch:
\[
\boxed{y(x) \sim x^{-1/4} \exp\left(\frac{2 x^{3/2}}{3\epsilon}\right)}
\]
\end{tcolorbox}

\subsubsection{Asymptotic series problems} \label{A:Asymptotic Series}
We described three types of problems in Section~\ref{sec:dataset} that were the hardest for LLMs to solve. In addition to these three problem types, we included other kinds of problems that LLMs found challenging. 

First, we included integrals $I(x)$ which can be approximated by an asymptotic series in the limit $x\to x_0$, for some fixed $x_0\in\mathbb{R}\cup\{\pm\infty\}$. We find a series $\sum_{n=0}^\infty a_n(x-x_0)^n$ such that $$I(x)-\sum_{n=0}^N a_n(x-x_0)^n << (x-x_0)^N$$ in the limit $x\to x_0$,  for $N$ fixed, though we do not require the difference to converge as $N\to\infty$. These asymptotic formulas often represent expansions around essential singularities. For example, consider the integral $$I(x)=\int^\infty_0\frac{1}{1+x^2t}{e^{-t}}dt$$ in the limit $x\to 0$. Using integration by parts to expand the integral we find that
$\int_0^\infty \frac{1}{1+x^2t}e^{-t}dt\approx\sum_{n=0}^\infty (-1)^nn!x^{2n}$, where we can check that the right-hand side is an asymptotic series. 

See an example of this technique below.

\begin{tcolorbox}[breakable, colframe=blue!60!yellow, colback=blue!5!white, title=Sample Asymptotic Series Problem and Full Solution]
\textbf{Problem:} 
Write the first two terms of the asymptotic series expansion of  \[I(x) = \int^x_1 \ln(xt^2)\cos(t^3) \mathrm{d}t\] in the limit $x \rightarrow \infty$. 

\thinline

\textbf{Solution:}
We will develop an asymptotic series using integration by parts.
Define 
\[u = \frac{\ln(xt^2)}{3t^2} \qquad \text{and} \qquad   v = \sin(t^3)\ .\]
Then
\[
\mathrm{d} u = \frac{-2(\ln(xt^2)-1)}{3t^3} \qquad \text{and} \qquad  \mathrm{d}v = 3t^2\cos(t^3)\ .\]
The formula \[\int u \mathrm{d}v = uv - \int v\mathrm{d}u\ .\]
gives us
\[I(x) = \left[\frac{\ln(xt^2)\sin(t^3)}{3t^2}\right]^x_1 + \int^x_1 \frac{2(\ln(xt^2)-1)\sin(t^3)}{3t^3} \mathrm{d}t\ .\]
We can apply integration by parts again to the remainder with
\[
u = \frac{2(\ln(xt^2)-1)}{9t^5}\qquad \text{and} \qquad v = -\cos(t^3)
\]
and their derivatives
\[
\mathrm{d}u = \frac{-2(5\ln(xt^2)-7)}{9t^6}\qquad \text{and} \qquad  \mathrm{d}v = 3t^2\sin(t^3)\ .
\]
Then we obtain

\[I(x) = \left[\frac{\ln(xt^2)\sin(t^3)}{3t^2} -\frac{2(\ln(xt^2)-1)\cos(t^3)}{9t^5}\right]^x_1 - \int^x_1 \frac{2(5\ln(xt^2)-7)\cos(t^3)}{9t^6} \mathrm{d}t\ .\]

Therefore, the first two terms of the asymptotic series expansion are
\[\boxed{ \frac{\ln(x^3)\sin(x^3)}{3x^2} - \frac{\ln(x)\sin(1)}{3} -\frac{2(\ln(x^3)-1)\cos(x^3)}{9x^5} + \frac{2(\ln(x)-1)\cos(1)}{9}}.\]
\end{tcolorbox}

\subsubsection{Integral problems}

In addition to asymptotic series problems, we include a broader class of one-dimensional parametric integrals of the form $I(\lambda) = \int^b_a \phi (\lambda; x) dt$, where the integrand $\phi (\lambda; x)$ may involve elementary functions, special functions, parameter-dependent exponents, singularities, polynomial or rational prefactors, or oscillations. These integrals are parameterized by a variable $\lambda$ that controls the problem's asymptotic behavior.

We are particularly interested in Laplace-type integrals with the form
$$ I(\lambda) = \int_a^b f(x) e^{-\lambda g(x)} dx
$$
We wish to construct analytical estimates $\phi_n(\lambda)$ of $I(\lambda)$ such that for each $n$, $\phi_n(\lambda)$ behaves similar to $I(\lambda)$ as $\lambda \to \infty$ for each $n$, and then find the optimal $N$ with estimate $\phi_N$. 

We first provide an example of an integral that can be solved using an ordinary taylor approximation.
\begin{tcolorbox}[breakable, colframe=blue!60!yellow, colback=blue!5!white, title=Sample Integral Problem and Full Solution]
\textbf{Problem:} 

Find the leading behavior up to $O(x^6)$ as $x \to 0_+$ of $$I(x)=\int_x^{1}\cos(xt)dt.$$ (Problem taken from \cite{bender2013advanced}.)

\thinline

\textbf{Solution:}

In finding the leading behavior of an integral

$I(x)=\int_a^b f(t,x) dt$ as $x \to x_0$

If $f(t,x) \sim f_0(t), x \to x_0$ uniformly in the interval $a \leq t \leq b$, then

$I(x)=\int_a^b f(t,x) dt \sim \int_a^b f_0(t) dt$ as $x \to x_0$

The function $\cos(xt)$ can be approximated for small $x$ with a Taylor series expansion

$$\cos(xt) = 1-\frac{(xt)^2}{2!}
            +\frac{(xt)^4}{4!}
            -\frac{(xt)^6}{6!}\dots$$

This series converges uniformly for $0 \leq x \leq t \leq 1 $, so we can integrate the terms in the Taylor series expansion in order to determine the leading behavior of this integral.

$$\int_x^{1}\cos(xt)dt \sim \int_x^{1} (1-\frac{(xt)^2}{2!}
            +\frac{(xt)^4}{4!}
            -\frac{(xt)^6}{6!}...)$$

$$= (1-x)-\frac{1}{2}x^2\left(\frac{1}{3} - \frac{x^3}{3}\right)
         + \frac{1}{24}x^4\left(\frac{1}{5} - \frac{x^5}{5}\right)
         - \frac{1}{720}x^6\left(\frac{1}{7} - \frac{x^7}{7}\right)...$$

The leading order behavior as $x \to 0+$ for the integral is given by

$$I(x) = (1-x)-\frac{1}{2}x^2\left(\frac{1}{3} - \frac{x^3}{3}\right)+ \frac{1}{24}x^4\left(\frac{1}{5} - \frac{x^5}{5}\right)- \frac{1}{720}x^6\left(\frac{1}{7} - \frac{x^7}{7}\right)$$

The solution up to order 6

$$\boxed{I(x) = 1 - x - \frac{x^2}{6} + \frac{x^4}{120} + \frac{x^5}{6}  - \frac{x^6}{5040} }$$

\end{tcolorbox}

We now provide an example applying Laplace's method to solve an integral.

\begin{tcolorbox}[breakable, colframe=blue!60!yellow, colback=blue!5!white, title=Sample Integral Ratio Problem and Asymptotic Solution]
\textbf{Problem:}

Estimate the leading-order behavior as \( x \to \infty \) of the ratio
\[
\left|\frac{\int_0^\infty \frac{t^{x-1}}{t+x}e^{-t^{1/4}}\,dt}{\int_0^\infty t^{x-1}e^{-t^{1/4}}\,dt}\right|
\]

\thinline

\textbf{Solution:}

To understand the behavior of the integrals, plot the function \( t^{x-1} e^{-t^{1/4}} \) for large \( x \); it becomes sharply peaked around a point \( t^* \). This localization allows us to approximate the slowly varying factor \( \frac{1}{t+x} \approx \frac{1}{t^* + x} \), and pull it out of the numerator integral:
\[
\left| \frac{\int_0^\infty \frac{t^{x-1}}{t+x}e^{-t^{1/4}}\,dt}{\int_0^\infty t^{x-1}e^{-t^{1/4}}\,dt} \right| \approx \left| \frac{1}{t^* + x} \cdot \frac{\int_0^\infty t^{x-1} e^{-t^{1/4}}\,dt}{\int_0^\infty t^{x-1} e^{-t^{1/4}}\,dt} \right| = \left| \frac{1}{t^* + x} \right|
\]

To locate the peak, define:
\[
\phi(t) = \ln(t^{x-1}) - t^{1/4} = (x-1)\ln t - t^{1/4}
\]
and solve \( \phi'(t^*) = 0 \) for the maximum:
\[
\phi'(t) = \frac{x-1}{t} - \frac{1}{4} t^{-3/4} \quad \Rightarrow \quad \frac{x-1}{t^*} = \frac{1}{4}(t^*)^{-3/4}
\]
Multiplying both sides by \( t^* \):
\[
x - 1 = \frac{1}{4} (t^*)^{1/4} \quad \Rightarrow \quad t^* = \left(4(x-1)\right)^4 \approx (4x)^4
\]

Thus:
\[
\left| \frac{1}{t^* + x} \right| \approx \frac{1}{(4x)^4} = \frac{1}{256\,x^4}
\]

Final Answer:
\[
\boxed{ \left|\frac{\int_0^\infty \frac{t^{x-1}}{t+x}e^{-t^{1/4}}\,dt}{\int_0^\infty t^{x-1}e^{-t^{1/4}}\,dt} \right| \sim \frac{1}{256\,x^4} \quad \text{as } x \to \infty }
\]
\end{tcolorbox}

\subsection{Evaluation setup}\label{A:Evaluation}
\subsubsection{Preparation of problem and solution}\label{A:Verification}

\begin{figure}[h]
    \centering
    \includegraphics[width=0.5\linewidth]{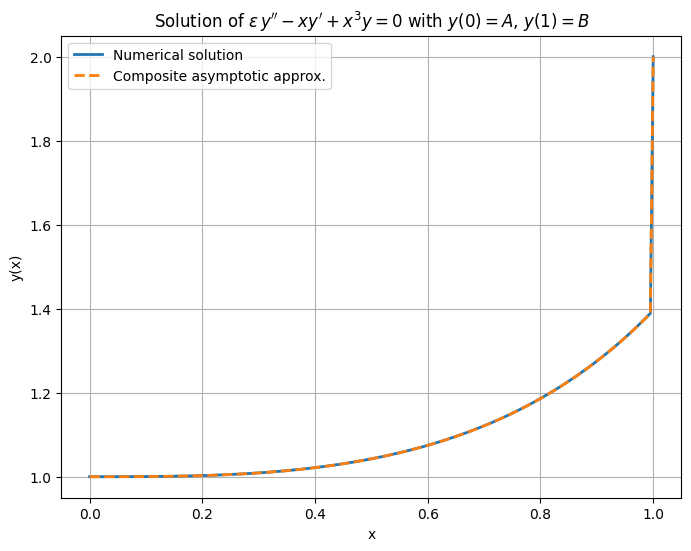}
    \caption{Visual comparison of numerical and approximate analytical solutions to a sample boundary value problem for solution verification.}
    \label{fig:visual}
\end{figure}

Each problem in the dataset went through a thorough verification procedure. For each problem generated by one student, another student was instructed to work through the same problem and ensure their answer matched the original solution. Figure~\ref{fig:visual} provides a visual comparison of the numerical and approximate solutions for the boundary value problem in Appendix~\ref{A:Boundary Layer}. This allows for semi-automated human verification that analytical solutions correspond well with numerical ground-truths, a method used to verify the problems in the dataset. Finally, students made sure that all of the problem statements and solutions followed consistent formatting guidelines that could be easily parsed and compared to the model responses.

\subsubsection{Prompts for response generation}\label{A:Prompts}
To ensure that model responses to the questions weren't driven by particular question wordings, we aimed to standardize prompts as much as possible based on the question type. Note that for the nonlinear PDE questions, there was some variance in the question types and so a representative task instruction (for when a problem asked for a self-similarity solution) is included below. If the nonlinear PDE questions additionally involved initial or boundary conditions, those were incorporated into the problem statement.

\begin{table}[!htb]
    \centering
    \caption{Prompts by Question Type}
    \begin{tabular}{p{3cm} p{3cm} p{6cm}}
        \toprule
        \textbf{Question Type} & \textbf{Inputs} & \textbf{Task instruction} \\
        \midrule
        WKB Approximation  & Problem, Initial Conditions & Find the leading order WKB approximation for the specific differential equation \verb|{Problem}| with initial conditions at \verb|{Initial Conditions}| where $\epsilon$ is a small positive parameter ($0 < \epsilon \ll 1$). Use only the variables and constants given in the problem; do not define additional constants. Place your final solution in a \verb|\boxed{}| LaTeX environment. \\
        \midrule
        Integral  & Problem, Limit & Consider the following integral: \verb|{Problem}| In the limit \verb|{Limit}|, find approximate behavior of the integral up to leading non-zero order in $\epsilon$. Provide your answer in a \verb|{\boxed{}}| LaTeX environment. \\
        \midrule
        Nonlinear ODE  & Problem, Limit & Find the leading order behavior of \verb|{Problem}| in the limit \verb|{Limit}|. Please place your final solution in a \verb|{\boxed{}}| LaTeX environment. \\
        \midrule
        Boundary Layer   & Problem, Boundary Conditions & Find a uniformly valid approximation to the solution of \verb|{Problem}| with boundary conditions \verb|{Boundary Conditions}| in the limit $\epsilon \ll 1$. Use only the variables and constants given in the problem; do not define additional constants. Place your final solution in a \verb|\boxed{}| LaTeX environment. \\
         \midrule
        Nonlinear PDE   & Problem, Limit & Find a self similarity solution for the non-linear partial differential equation \verb|{Problem}| in the limit \verb|{Limit}|. Please place your final solution in a \verb|\boxed{}| LaTeX Environment. \\
         \midrule
        Asymptotic Series   & Problem, Limit & Find the first two terms in the asymptotic series of \verb|{Problem}| in the limit \verb|{Limit}|. Provide your answer in a \verb|\boxed{}| LaTeX environment. \\
        \bottomrule
    \end{tabular}
    \label{tab:task_instructions}
\end{table}

The prompts were combined with the following prompt suffix: Place your final answer in a \verb|\boxed{}| LaTeX environment. If you have multiple answers, separate them with a ``;". Use the notation from the problem and do not define any new variables.

\subsubsection{Numerical evaluation}\label{A:Numerical}

Part of the importance of having a standardized prompt and response format was to ensure consistency in numerical evaluation of the solutions provided by students and the various models. Provided \verb|\boxed{}| LaTeX solutions were parsed into symbolic representations in a method discussed in \ref{A:Parser}. These symbolic representations were then evaluated at a particular value for all of the input variables into the problem. For example, if a problem had its solution in terms of the variable $x$, the solution was evaluated at a particular value for $x$. The model solutions were then graded whether at that value they numerically matched the verified student solutions.

\subsection{Automated parsing and model evaluation}\label{A:Parser}

To compare LLM-generated solutions against ground-truth solutions written by students, our parser converts LaTeX expressions into symbolic representations that can be programmatically evaluated for numerical closeness.

The parser architecture is designed to handle complexities and variations in mathematical notation. Initially, the system extracts solutions from LaTeX \texttt{\textbackslash boxed\{\}} environments using regular expression pattern matching. It also is able to process multiple solutions using the semicolon as a delimiter. This extraction ensures that only the final answers are evaluated, filtering out intermediate text. After the extraction, the system converts LaTeX notation into SymPy expressions through a series of transformation rules, such as replacing Unicode characters with LaTeX code and removing unnecessary formatting. The parser then processes specialized symbols like integrals, beta functions, and expressions with superscripts and subscripts to translate them into a standardized format.

All models are prompted to provide their final answer in a LaTeX \texttt{\textbackslash boxed\{\}} environment. Then, a custom-built parser uses Python's RegEx library to search for the final solution at the end of the model's output. The parser then converts the extracted LaTeX expression into a SymPy expression. In our types of applied math problems, it is possible for different functional forms to simultaneously be valid approximations; therefore, we check correctness by   evaluating the model's SymPy expression and the ground-truth solution produced by students---which is also converted into a SymPy expression---at the same point in the domain and determining whether the values are within a closeness threshold. This approach to evaluation ensures that the model's solution is considered correct only if it precisely matches the ground-truth solution. We ensure that there is no ambiguity in the prompts to the model by specifying any variables or free parameters---therefore, all problems can be answered as a function only of their independent variables.

\subsection{Additional analysis of model failure modes}\label{Sec:model_analysis}

Given that our evaluation framework strictly required solutions in a \verb|\boxed{}| environment, we noticed that some models failed to follow these instructions. In particular, DeepSeek-R1 never placed its final solution in a LaTeX box. Our first instinct was that the model was exceeding its max tokens, which is set by default to 32,768 tokens. However, we found that the model was rarely reaching the limit and simply failed to converge to a final answer written in proper mathematical formatting. The model also often replaced mathematical expressions with LaTeX ellipses (specifically, $\dots$), presumably for convenience, but never plugged the true mathematical symbols or expressions back in to its final result. Therefore, we did not include the model in our evaluation results.

Other models also exhibited difficulty following the instructions in our prompts. In particular, despite its high performance on our benchmark, Gemini 2.5 Pro frequently ignored formatting guidelines, using different LaTeX formats to flag its final solution. This is one potential explanation for why Gemini 2.5 Flash Thinking showed higher overall accuracy than Gemini 2.5 Pro, despite the latter supposedly being Google's most performant model.

\newpage

%%%%%%%%%%%%%%%%%%%%%%%%%%%%%%%%%%%%%%%%%%%%%%%%%%%%%%%%%%%%

\end{document}